\documentclass[sigconf]{acmart}

\usepackage{graphicx} 
\usepackage{enumitem}
\usepackage{hyperref}
\usepackage[title]{appendix}
\usepackage{listings}
\usepackage{pifont}
\usepackage[labelfont=bf]{caption}
\usepackage{makecell}
\usepackage{stfloats}
\usepackage[hyphenbreaks]{breakurl}
\usepackage{float}
\usepackage{tabularray}
\usepackage{longtable}

\usepackage{color}

\definecolor{dkgreen}{rgb}{0,0.6,0}
\definecolor{gray}{rgb}{0.5,0.5,0.5}
\definecolor{mauve}{rgb}{0.58,0,0.82}

\setcopyright{none}

\title[Towards Grammatical Tagging for the Legal Language of Cybersecurity]{Towards Grammatical Tagging\\ for the Legal Language of Cybersecurity}
\author{Gianpietro Castiglione}
\email{gianpietro.castiglione@phd.unict.it}
\affiliation{
\institution{University of Catania, Department of Mathematics and Computer Science}
\country{Italy}
}

\author{Giampaolo Bella}
\email{giamp@dmi.unict.it}
\affiliation{
\institution{University of Catania, Department of Mathematics and Computer Science}
\country{Italy}
}

\author{Daniele Francesco Santamaria}
\email{daniele.santamaria@unict.it}
\affiliation{
\institution{University of Catania, Department of Mathematics and Computer Science}
\country{Italy}
}

\copyrightyear{2023}
\acmYear{2023}
\setcopyright{acmlicensed}\acmConference[ARES 2023]{The 18th International Conference on Availability, Reliability and Security}{August 29-September 1, 2023}{Benevento, Italy}
\acmBooktitle{The 18th International Conference on Availability, Reliability and Security (ARES 2023), August 29-September 1, 2023, Benevento, Italy}
\acmPrice{15.00}
\acmDOI{10.1145/3600160.3605069}
\acmISBN{979-8-4007-0772-8/23/08}

\begin{document}

\begin{abstract}
Legal language can be understood as the language typically used by those engaged in the legal profession and, as such, it may come both in spoken or written form. Recent legislation on cybersecurity obviously uses legal language in writing, thus inheriting all its interpretative complications due to the typical abundance of cases and sub-cases as well as to the general richness in detail. This paper faces the challenge of the essential interpretation of the legal language of cybersecurity, namely of the extraction of the essential Parts of Speech (POS) from the legal documents concerning cybersecurity. 

The challenge is overcome by our methodology for POS tagging of legal language. It leverages state-of-the-art open-source tools for Natural Language Processing (NLP) as well as manual analysis to validate the outcomes of the tools. As a result, the methodology is automated and, arguably, general for any legal language following minor tailoring of the preprocessing step. It is demonstrated over the most relevant EU legislation on cybersecurity, namely on the NIS 2 directive, producing the first, albeit essential, structured interpretation of such a relevant document. Moreover, our findings indicate that tools such as SpaCy and ClausIE reach their limits over the legal language of the NIS 2.
\end{abstract}


\maketitle

\section{Introduction}\label{sec:intro}
It is widely accepted that, in its attempt to fully describe all applicable circumstances and details, legal language may become lengthy and complicated, hence various professional figures specialise precisely in the interpretation of the contents of documents written in such a language. Legal language is, of course, also used at the level of European legislation, which promotes several types of legal acts, namely Regulations, Directives, Decisions, Recommendations and Opinions~\cite{legislation}.

European legislation also touches upon cybersecurity, notably through the ``Directive (EU) 2022/2555 of the European Parliament and of the Council of 14 December 2022 on measures for a high common level of cybersecurity across the Union'', known as NIS 2 in brief~\cite{nis2}, which can be considered the most relevant treatment of cybersecurity in legal language at European level today.
We observe that lengthy written information may complicate, in general, even the essential interpretation step known as \textit{grammatical tagging} or \textit{POS tagging}, which consists of extracting the main components of the discourse, such as clauses of a sentence and subjects of a clause. This might be problematic not only for professionals such as lawyers but also for the layman who is called to read and accept various terms of service on the web several times a day.

In light of these observations, we want to investigate how to use computer-support to assist through the POS tagging of legal language.
The present work leverages NLP techniques to develop a methodology offering machine-assisted support through the extraction of the Parts Of Speech (POS) of legal language. In consequence, the understanding of directives and similar documents does no longer have to be entirely manual but can be automated by applying our methodology. More precisely, the task our methodology automates is information extraction for POS tagging, which consists in the identification of the main agents and of the actions they must perform according to the given text. 

Therefore, our methodology seeks out to extract at least the relevant clauses from each sentence as well as to identify the relevant POS from each clause, namely the subject (agent), the main verb or predicate (action), and the object of the action (object). Because we aim at open science, our work leverages the forefront of open-source techniques, namely the SpaCy and ClausIE libraries, as we shall see below. We understood from our early experiments that, once tailored to legal language, such libraries reach their limits, hence our methodology requires the analyst to validate their outputs by a systematic comparison with a manual analysis of sentences.

Our methodology is arguably general, hence it can be applied to any legal document. However, we specifically demonstrate it on the legal language of modern cybersecurity directives, in particular on the NIS 2 Directive, which can be considered the best benchmark. We analysed its 46 articles and selected those where actual prescripts are made, leaving away those about general prerequisites and terminology. As a result, we grammatically analysed articles 7 through to 37 by both a parser based on ClausIE and manually, concluding that manual validation cannot be avoided over legal language. Precisely, we found that the tools are good in tagging subjects on average on 96.1\% of cases, then verbs on 83.3\% and objects only on 64.5\% of cases.

An overview of the related work and of SpaCy are in Section \ref{sec:related} and in Section \ref{sec:outline} respectively. The method for grammatical tagging of legal language is illustrated step by step in Section \ref{sec:meth} while its application to NIS 2 Directive is in Section \ref{sec:meth-dem}. The statistical analysis is evaluated in Section \ref{sec:stat} and the open challenges encountered are highlighted in Section \ref{sec:open-ch}. In Section \ref{sec:concl} the conclusions are outlined.

\section{Related work}\label{sec:related}
Several works have been conducted in the legal field by leveraging NLP and machine learning algorithms. 
The new capabilities made available by artificial intelligence have even allowed the prediction of judgements, the search for inconsistencies between the various legal processes and other tools for the simplification of legal context.
This Section only outlines the most recent and most closely related to the contributions of this paper.

The LEGAL-BERT tool adapts the BERT models to the legal context; the main feature is to predict some deliberately maskable words related to the legal context~\cite{chalkidis-etal-2020-legal}. 
The scope of the prediction is even wider; Medvedeva et al. investigated the use of NLP for predicting legal decisions, and in particular for predicting if there were  any violations of the European Convention on Human Rights~\cite{Medvedeva2020}. At the same time, Katz et al. constructed a model, based on the random forest approach, for predicting, using only available data, years and years of decisions by the Supreme Court of the United States~\cite{katz2014predicting}.

Particular attention has been given to the use of NLP in the context of patents: Arts et al. use ``natural language processing techniques to harness the rich content of patent documents, identify new technologies and their impact\ldots validation studies support the use of text mining techniques to identify new technologies and measure patent novelty at the time of filing, and to measure the impact of these new technologies on subsequent innovation.''~\cite{ARTS2021104144}; while Sheng Lee et al. developed a deep learning pre-trained model for the generation of patent claims~\cite{lee2019patent}.

NLP techniques have been not used only for prediction purposes and for text generation but also for questioning and answering. 
Zhong et al. presented a question-answering data-set for the legal domain that collects questions from the National Judicial Examination of China ~\cite{zhong2019jecqa}. They reason on the data-set for facing the challenges of word matching, concept understanding or multi-paragraph reading. 
Similar classifications for legal documents were built by de Araujo et al. with the additional feature of leveraging NLP for theme assignment and labelling, regarding Brazilian Supreme Court \cite{luz-de-araujo-etal-2020-victor}, and by Sulea et al., which apply machine learning techniques for predicting and investigating on the sentences of French Supreme Court ~\cite{sulea2017exploring}.

Artificial intelligence techniques have been used also to solve particular cases of inconsistencies. Xu et al. developed a framework for legal judgment prediction) tasks ~\cite{xu-etal-2020-distinguish}. The goal of the framework is to face confusing charges by analysing similar articles and fact descriptions, thanks to a new graph distillation operator.
Ul Hassan et al. leverage NLP and machine learning algorithms for legal text classification \cite{doi:10.1061/(ASCE)LA.1943-4170.0000379}. They introduced new models for identifying and extracting, from construction contracts, the two categories of requirements and non-requirements.
Lippi et al. face the cases of terms of service of online platforms. In particular, they leveraged machine learning for detecting if any unfair clauses for the user of the online platform exist \cite{Lippi_2019}.

In the mere context of information extraction two main branches can be identified: POS tagging and topic modelling. In the first case, several libraries for different programming languages exist, for example, Scikit-learn ~\cite{scikit-learn}; in the second case, Latent Dirichlet Allocation (LDA) is one of the most used approaches ~\cite{jelodar2018latent}. LDA, introduced by Blei, Ng and Jordan ~\cite{10.5555/944919.944937}, is a generative probabilistic model and represents documents as random mixtures over several topics, where a topic is produced by distribution among different words.

It seems fair to claim, from the summary given above, that the main works are based on ad-hoc development of NLP or machine learning algorithms within a specific use case. 
In particular, the main techniques aim at the prediction of text and situations or, alternatively, at classification as their main goals. 
Such approaches are certainly valid but bear the inherent limitations of being heavily dependent on pre-trained models related to the specific use case, which in turn require the availability of huge amounts of labelled data and consequently high-performance hardware.

\section{An outline of SpaCy} \label{sec:outline}
Following the related work just summarised, it is clear that POS tagging is only a specific niche of NLP and that, in particular, no significant attempts to apply POS tagging to legal language are worth noting. Following these observations, we aim at approaches that are lightweight and, at the same time, general so as to be applicable to virtually any legal language. 
In the context of the work presented in this paper, we shall use what is perhaps the best-established tool for processing text, namely the SpaCy library ~\cite{spacy2}, whose main peculiarity beyond POS is Named Entity Recognition. 

SpaCy is a free library for the Python programming language and can be considered among the most versatile tools to tailor NLP to practical applications. In order to get the correct token, SpaCy assigns syntactic dependency labels that allow us to identify subjects, verbs or objects.
SpaCy both works on pre-trained pipelines related to a specific language hence context-independent, as well as on local data.
The SpaCy pipeline of actions is essentially as follows: first, the words in input are tokenised. After that, each token is tagged to a specific part of speech, leveraging the pre-trained models, thereby predicting the closest role of each token derived from the input.

In practice, it is very useful to appeal to a sub-library of SpaCy called ClausIE (claucy) ~\cite{claucy}, which is able to divide the input text into the main components of each statement, namely into clauses. We shall see below several examples supporting the case that the complexity of a legal text can be daunting even for such powerful, state-of-the-art NLP tools. Moreover, we are aware that POS tagging is only one step towards automated information extraction from such a language, while its semantic interpretation can be expected to be even more challenging.

\section{A method for grammatical tagging of legal language} \label{sec:meth}
Our method for the grammatical tagging of any document written in legal language follows a waterfall style and is completed with manual validation, as we shall see below. In short, given a legal document, it takes as input some text written in a legal language and carries out a punctual analysis of the input in order to output the POS tagging of each part of the text.

\paragraph{Step 1. Preprocessing}
In order for our method to also automate document parsing, a preliminary preprocessing phase is necessary. Our target, in fact, is not just information extraction but includes automating the selection of each sentence subjected to tagging.
Since each legal document has its own structure, the parsing must necessarily be manually preset and context-dependent.
In general, we want to accomplish the following steps:

\paragraph{a) Identification of articles}
Assuming we are analysing the legal language of a document that is composed of articles, this step consists in identifying the more significant ones where we can extract relevant POS. These are normally those articles where the actions to be carried out by the entities defined in the document itself are defined in precise detail. By contrast, our target articles do not include those with abbreviations and general provisos. It follows that the correct articles can only be determined with a manual and subjective selection of the range to be considered. In fact, if we parse all the articles without distinction, we could get an inconsistent collection of the POS, because some might not represent the notions of actions we aim to distil.

\paragraph{b) Text acquisition}
After finding the range of articles to examine, this step consists of acquiring all the text from a given Article with the aim of automated processing. We shall expect a subdivision into items, which can be numeric or alphabetical, within each article, a style that is quite common in legal documents.
A useful acquisition heuristic is to select all text between the string ``Article \emph{i}'' and ``Article \emph{i+1}'' in order to collect all text of the $i$-th article, of course having checked that such keywords do not occur in the text forming the body of the article.
Moreover, if the selection involves two different pages, then spurious information must be deleted.

\paragraph{c) Item selection}
To input a text, such as a legal one, to any NLP technique, it is necessary to define a heuristic for the choice of the  \textit{single units of processing} inside the given text, namely the fragments of text that, given as a whole to the chosen technique, leads to producing new, relevant information. By contrast, applying it to a whole article, which may consist of several sentences, may not lead to useful information. 
As our aim is to identify single subjects, actions and objects, our single unit of processing is represented by each single sentence. 

\paragraph{d) Hierarchy identification}
Sometimes, the single unit of processing may have to be built by copying parts of the statement, for example, subject and action, which may have been omitted to limit redundancy. This is typically the case when an item from an itemised list is specialised by a whole itemised sub-list. In such a typical case, this step associates the underlying item with the overlying one, namely by removing the colon and linking the underlying hierarchical level with the one above. However, it is necessary to manually verify the unit of processing that is built this way so as to minimise the risk of inconsistencies. 

\paragraph{Step 2. POS tagging}
As mentioned, grammatical tagging of a sentence extracts the parts of speech and, first of all, identifies the clauses from each unit of processing.
Once clauses are identified, each clause may contain up 5 different sentence patterns: Subject (S), Verb (V), Complement (C), Object (C), Adverb (A). In particular, the difference between C and O relies on the role of each of them: a Complement completes the meaning of a sentence while Object is specifically the direct object on which the action of the verb is reflected.
Of course, while only one S and V pair should be expected per clause, several C and O instances may be encountered.

For example, patterns can be extracted by the code snipped Code~\ref{code:1}, which leverages our chosen libraries.
It can be seen that, firstly, it is necessary to load the pre-trained model, and the choice of \emph{en\_core\_news\_lg} means adopting the largest model consisting of 514k keys for a total of 560 MB. 
In all steps of the method, this represents the highest computational cost but it is arguably necessary to deal with as many cases as possible.
The pre-trained model is then loaded into the pipeline and the ``engine'' for text pattern extraction is ready. 
The code returns the extracted clauses together with all sentence patterns.

\begin{lstlisting}[language=python, caption=Python code snippet for the extraction of sentence clauses and patterns based on SpaCy and ClausIE, label={code:1}]
import spacy
import claucy

nlp = spacy.load("en_core_news_lg")
claucy.add_to_pipe(nlp)
    
def NLP(text):
    document = nlp(text)
    return document._.clauses, document._.clauses[0].to_propositions(as_text=True)
\end{lstlisting}

The same thing can be programmed in a different way, as illustrated in Code~\ref{code:2} \cite{packt}.
The main difference is that Code \ref{code:2} leverages SpaCy only. For the object extraction, in fact, the dependency tree generated by the tagging step will be parsed. Only when the tag related to the object is obtained, the sub-text identified by the related sub-tree will be returned. This can be limiting and less precise than using ClausIE, since the latter uses a search based on sentences rather than tags. Our preliminary experiments support this claim, in fact, it could happen that the objects identified by ClausIE would not be identified by SpaCy alone, indicating that our first approach is preferable to the second, while performance difference is negligible.

\begin{lstlisting}[language=python, caption=Python code snippet for the extraction of sentence clauses and patterns based on Spacy only, label={code:2}]
def get_object_phrase(doc):
    for token in doc:
        if ("dobj" in token.dep_):
            stree = list(token.subtree)
            a = stree[0].i
            b = stree[-1].i + 1
            return doc[a:b]
\end{lstlisting}

\paragraph{Step 3. Tabulation} 
For a correct statistical analysis of the validity of the method, relevant data must be collected and structured.
The template that will be filled in for each single article is shown in Table \ref{tab:tab-template}.
For each relevant element we extracted by NLP functions in the previous step we dedicate three columns, for a total of 10 columns considering an additional one for the identification of the single item. 
In particular, the first column features the correct POS that we extracted by a manual analysis of the sentence, then the second column states the POS extracted by our use of the NLP libraries, while the third column indicates an evaluation of the correctness of the automated extraction with respect to our manual one, using an alphabet of four symbols, represented in Table \ref{tab:legend-sym}. In particular, a tick symbol means that the two analyses coincide, then a cross and a vertical bar symbol respectively mean that we found the automated analysis to be entirely wrong or partially wrong. Additionally, a ``P'' symbol stands for clauses in passive style.

\begin{table}[h]
\caption{Legend of symbols}
    \centering
    
    \begin{tabular}{|c|c|}
      \hline
      \textbf{Symbol} & \textbf{Acronym}\\
      \hline
      \ding{52} & Correct answer \\
      \hline 
      \ding{56} & Wrong answer \\
      \hline 
      \ding{121} & Correct but incomplete answer \\
      \hline 
      P & Passive Verb \\
      \hline
  \end{tabular}
  
  \label{tab:legend-sym}
\end{table}

\begin{table*}[t]
\caption{Table template for grammatical tagging of an article}
  \begin{tabular}{|l||c|c|c||c|c|c||c|c|c|}
    \hline
    N. & Sub & I-Sub & Sub-HIT & Verb & I-Verb & Verb-HIT & Obj & I-Obj & Obj-HIT \\
    \hline
            \dots & \dots & \dots & \dots & \dots & \dots & \dots & \dots & \dots & \dots
    \\ \hline
    & \multicolumn{3}{c||}{\textbf{Sub-HIT \% =  \%}} &
    \multicolumn{3}{c||}{\textbf{Sub-HIT \% =  \%}} & 
    \multicolumn{3}{c|}{\textbf{Obj-HIT \% =  \%}} \\
    \hline
\end{tabular}
\label{tab:tab-template}
\end{table*}

The last row of Table \ref{tab:tab-template} refers to the percentage of validity of the solutions found by the automated approach. 
It could be calculated by a simple metric, for example by assigning full value to the ticks, no value to the crosses, half value to the bars and no value to passive clauses. 

\section{Demonstrating our method on the NIS 2 Directive} \label{sec:meth-dem}
This section illustrates the application of our method to the NIS 2 Directive. For the sake of demonstration, the method is illustrated on Article 11 and Article 23 because they cover different cases.
For each step, the following extracts will be analysed:

\vspace*{1\baselineskip}

\fbox{\Small
\parbox{0.44\textwidth}{
    2. Member States shall ensure that their CSIRTs jointly have the technical capabilities necessary to carry out the tasks referred to in paragraph 3. Member States shall ensure that sufficient resources are allocated to their CSIRTs to ensure adequate staffing levels for the purpose of enabling the CSIRTs to develop their technical capabilities.

    \dots

    8. At the request of the CSIRT or the competent authority, the single point of contact shall forward notifications received pursuant to paragraph 1 to the single points of contact of other affected Member States.
    }
}
\paragraph{Step 1. Preprocesing}
By coherently applying the various sub-steps, we have the following outcomes:

\paragraph{a) Identification of articles}
In the NIS 2 Directive, the articles are numerated  from 1 to 46. However, not all of them are significant from the point of view of the measures. In fact, all articles until 7 are preliminary considerations and definitions, while from the 38 onward various considerations on the applicability of the Directive are described.
Therefore, we will parse the Directive from Article 7 to Article 37.

\paragraph{b) Text acquisition}
To acquire the text of Article \emph{i}, the more efficient solution in the context of NIS 2 Directive is as anticipated above, namely to grab the text between each pair of article headers of the form ``Article \emph{i}'' and ``Article \emph{i+1}'' as strings. Moreover, we must make sure to expand each header with two characters of line feed both at the start and at the end 
because some articles could be cited through some items both by the numerical notation and by their full title, a scenario that would hinder the parsing.
By contrast, stating which and where the line feed characters are ensures that we capture an occurrence at the beginning of an article, namely an actual header.

An emblematic case in which the described heuristic is necessary is represented by the following extract:
\vspace*{1\baselineskip}

\fbox{\Small
  \parbox{0.44\textwidth}{
    \begin{center}
Article 10

\textbf{Computer security incident response teams (CSIRTs)}
\end{center}
\vspace*{2\baselineskip}
1. Each Member State shall designate or establish one or more CSIRTs. The CSIRTs may be designated or established within a competent authority. The CSIRTs shall comply with the requirements set out in Article 11(1), shall cover at least the sectors, subsectors and types of entity referred to in Annexes I and II, and shall be responsible for incident handling in
accordance with a well-defined process. 

\dots

10. Member States may request the assistance of ENISA in developing their CSIRTs.
\vspace*{2\baselineskip}
\begin{center}
Article 11

\textbf{Requirements, technical capabilities and tasks of CSIRTs}
\end{center}
  }%
}

\paragraph{c) Item selection} 
Coherently with the definition of this step, a single sentence is selected by referring to its termination by means of the full stop, in the simplest case. The errors that could arise through this approach concern the case in which the full stop does not terminate the sentence. This case only happens in the case of hierarchical lists and is rather frequent through the NIS 2. It will be treated below. 

\paragraph{d) Hierarchy identification}
In the NIS 2 Directive, each item could have up to two hierarchical levels of depth. The first is listed by letters, while the second is listed by ordinal numbers. Developing a parser that identifies such cases is not complex, since bulleted lists can be used as selection criteria. By contrast, it may be difficult to understand what is represented at the lowest hierarchical level.
In the simplest case, we can resolve the following form of compression (Article 9, item 4):
\begin{quote}
That plan shall lay down, in particular:
    (a) the objectives of national preparedness measures and activities;
\end{quote}

by making the preliminary part of the statement explicit in all possible items, for example, as follows:

\begin{quote}
That plan shall lay down, in particular, the objectives of national preparedness measures and activities;
\end{quote}

In such a case, it generally happens that the primary sentence expresses the presence of some tasks that are described at the first hierarchical level. Consequently, the first hierarchical level states significant verbs that describe the specific actions of the defined subject. Therefore, the second hierarchical level aims at further completing the meaning, defining the objects that the action affects.

\paragraph{Step 2. POS tagging}
By executing Code \ref{code:1} on some sentences of the previous extracts, the following outputs are produced:
\begin{lstlisting}[caption=NLP output on Article 11, label={out:11}]
[<SVC, Member States, shall ensure, None, None, that their CSIRTs jointly have the technical capabilities necessary to carry out the tasks referred to in paragraph 3, []>, <SVO, their CSIRTs, have, None, the technical capabilities necessary to carry out the tasks referred to in paragraph 3, None, [jointly]>]
\end{lstlisting}

\begin{lstlisting}[caption=NLP output on Article 23, label={out:23}]
[<SV, the single point of contact, shall forward, None, None, None, []>, <SV, the single point of contact shall forward notifications, received, None, None, None, [At the request of the CSIRT or the competent authority, pursuant to paragraph 1, to the single points of contact of other affected Member States]>]
\end{lstlisting}

In the present work, the desired combinations of sentence patterns are SVO or SVC because in these combinations there are all the POS that are necessary for a subsequent semantic interpretation. 
If complex sentences are analysed, the combinations given as output may not be unique and definitive, as in the cases above, where two patterns were extracted for each article. Therefore, it would be necessary to manually choose the most appropriate pattern.

\begin{table*}[ht]
\centering \small
\caption{Table for grammatical tagging of NIS 2 Article 11}
  \begin{tabular}{|l||c|c|c||c|c|c||c|c|c|}
    \hline
    N. & Sub & I-Sub & Sub-HIT & Verb & I-Verb & Verb-HIT & Obj & I-Obj & Obj-HIT \\
    \hline 
    1 & - & - & - & - & - & - & - & - & - \\
    1.a & C & C & \ding{52} & ensure & ensure & \ding{52} & \ref{n11-1-1a} & \ref{s11-1-1a} & \ding{121} \\
    1.b & C Premises and SYS & C premises & \ding{121} & P - located & located & \ding{52} & \ref{n11-1-1b} & \ref{s11-1-1b} & \ding{56} \\
    1.c & C & C & \ding{52} & P - equipped & equipped & \ding{52} & \ref{n11-1-1c} & \ref{s11-1-1c} & \ding{56} \\
    1.d & C & C & \ding{52} & ensure & ensure & \ding{52} & \ref{n11-1-1d} & \ref{s11-1-1d} & \ding{52} \\
    1.e & C & C & \ding{52} & P - staffed & ensure |  ensure & \ding{52} & \ref{n11-1-1e} & \ref{s11-1-1e} & \ding{56} \\
    1.f & C & C & \ding{52} & P - equipped & equipped  & \ding{52} & \ref{n11-1-1f} & \ref{s11-1-1f} & \ding{56} \\
    \hline
    2.1 & MS & MS & \ding{52} & ensure & ensure & \ding{52} & \ref{n11-2-1} & \ref{s11-2-1} & \ding{52} \\ 
    2.2 & MS & MS & \ding{52} & ensure & ensure & \ding{52} & \ref{n11-2-2} & \ref{s11-2-2} & \ding{52} \\
    \hline
    \dots & \dots & \dots & \dots & \dots & \dots & \dots & \dots & \dots & \dots
    \\ \hline
    & \multicolumn{3}{c||}{\textbf{Sub-HIT \% = 95 \%}} &
    \multicolumn{3}{c||}{\textbf{Sub-HIT \% = 90 \%}} & 
    \multicolumn{3}{c|}{\textbf{Obj-HIT \% = 45 \%}} \\
    \hline
    \end{tabular}  
    \label{tab:tab-art-11}
\end{table*}

\begin{table*}[ht]
\centering \small
\caption{Table for grammatical tagging of NIS 2 Article 23}
\begin{tabular}{|l||c|c|c||c|c|c||c|c|c|}
    \hline
    N. & Sub & I-Sub & Sub-HIT & Verb & I-Verb & Verb-HIT & Obj & I-Obj & Obj-HIT \\
    \hline 
    1.1 & MS & MS & \ding{52} & ensure & ensure & \ding{52} & 
    \ref{n23-1-1} & \ref{s23-1-1} & \ding{52} \\
    1.2 & Entities concerned & Entities Concerned & \ding{52} & ensure & ensure & \ding{52} & \ref{n23-1-2} & \ref{s23-1-2} & \ding{52} \\
    1.3 & MS & MS & \ding{52} & ensure & ensure & \ding{52} & \ref{n23-1-3} & \ref{s23-1-3} & \ding{52} \\
    1.4 & Mere ... notification & Mere ... notification & \ding{52} & not subject & subject & \ding{56} & \ref{n23-1-4} & \ref{s23-1-4} & \ding{52} \\
    \hline 
    \dots & \dots & \dots & \dots & \dots & \dots & \dots & \dots & \dots & \dots \\
    \hline
    8 & POC & POC & \ding{52} & forward & NONE & \ding{56} & \ref{n23-8-1} & \ref{s23-8-1} & \ding{56} \\
    \hline
    9.1 & POC & POC & \ding{52} & submit & submit & \ding{52} & \ref{n23-9-1} & \ref{s23-9-1} & \ding{52} \\
    9.2 & E & E & \ding{52} & contribute & contribute & \ding{52} & \ref{n23-9-2} & \ref{s23-9-2} & \ding{52} \\
    9.3 & E & E & \ding{52} & inform & inform & \ding{52} & \ref{n23-9-3} & \ref{s23-9-3} & \ding{52} \\
    \hline
    10 & C & C & \ding{52} & provide & provide & \ding{52} & \ref{n23-10-1} & \ref{s23-10-1} & \ding{56} \\
    \hline
    11 & Commission & Commission & \ding{52} & adopt & adopt & \ding{52} & \ref{n23-11-1} & \ref{s23-11-1} & \ding{56}
    \\ \hline
    & \multicolumn{3}{c||}{\textbf{Sub-HIT \% = 92.8 \%}} &
    \multicolumn{3}{c||}{\textbf{Sub-HIT \% = 80.9 \%}} & 
    \multicolumn{3}{c|}{\textbf{Obj-HIT \% = 83.3 \%}} \\
    \hline
    \end{tabular}
    \label{tab:tab-art-23}
\end{table*}

In consequence, obtaining only pattern SV may be taken to signify incompleteness of the automated approach, while pattern SVOC falls into the cases that can be considered complete. 
The whole output consists of sentence pattern type, sentence subject, sentence verb, sentence indirect object, sentence direct object, sentence complement and sentence adverbials.
Our experiments indicated that all possible combinations of the 5 can be output on the NIS 2.

It can be seen from Code \ref{out:11} that two solutions were generated, both containing the object, but only the first solution is complete. In fact, the correct Subject was identified, together with the Verb and the Complement. Although the second identified sentence pattern might have seemed reasonable, the POS identified are not correct. 

Code \ref{out:23} shows that the produced sentence pattern SV indicates no object was identified. While we can see that both identified sentence patterns  produced the correct Subject, the second sentence pattern is wrong in the handling of the verb. This further confirms that the automated approach is not perfect.

A significant case that was encountered is the presence of different hierarchical levels with a specific feature: the object of the sentence introducing the hierarchical level explicitly expresses their presence. This leads to inconsistencies. For example: \emph{Member State shall have the following tasks: a) \dots b) \dots} .
There is greater complexity towards grammatical tagging in the presence of the second hierarchical level.

\paragraph{Step 3. Tabulation} 
Iterating the extraction over all relevant Articles allows us to collect all information we aim at. The POS of articles 11 and 23 are depicted respectively in Table \ref{tab:tab-art-11} and Table \ref{tab:tab-art-23}, while the used acronyms are defined in Table \ref{tab:legend-acr}. Our symbols purposely differentiate the incomplete answer from the incorrect answer, since the first one is to be considered partially correct. Therefore its extraction does not have to be considered entirely misleading.

\begin{table}[h]
\centering \small
\caption{Legend of acronyms}
\begin{tabular}{|c|c|} 
      \hline
      \textbf{Name} & \textbf{Acronym}\\
      \hline
      Member State & MS  \\
      \hline
      National Cybsersecurity Strategy & NCS \\ 
      \hline
      Competent Authority & CA \\ 
      \hline
      Commission & Co \\
      \hline
      Point of Contact & POC \\
      \hline
      CSIRT & C \\
      \hline
      ENISA & E \\
      \hline
      Cooperation Group & CG \\
      \hline
      The European External Action Service & EEAS \\
      \hline
      CN & CSIRTs Network \\
      \hline
      Eu-Cyclone & EuC \\
      \hline
      SYS & Supporting Information Systems \\
      \hline
    \end{tabular}  
  \label{tab:legend-acr}
\end{table}

The tables exhaustively list the occurrences of relevant extracted POS. The tables provide a clear and first-level view of the NLP results previously discussed. 
Naturally, more correct results are obtained where there is less complexity of the related POS, therefore in most of the cases, simplicity concerns the subjects.

\section{Statistical analysis} \label{sec:stat}
To evaluate the overall functioning of the automated tagging with respect to our manual one, 
we average the hit values for each tag of each grammatical tagging table.
As suggested above, we associate hit value 1 with a correct answer, 0.5 with an incomplete answer and 0 otherwise. 
Therefore, for each table \emph{i} and for each tag \emph{j}, the Hit Rate (HR) is the sum of the HIT values divided by the number of occurrences:
\[ HR_{i,j} = \frac{\sum\ HV_{i,j} }{\#occ_{i,j}} \cdot 100,\ \rm{where}\ 7 \le i \le 37, 1 \le j \le 3.\]

We decided to skip the following cases: measures with no item number, measures that may include a priori contextualisation and hierarchical levels introduced by ``have following tasks'' and similar ones where no other agent is expressed. 
The reasons are, respectively: difficulty in automatically parsing the measure since it has no identifier (for example, Article 17 and Article 36); information extraction is sometimes wrong because the libraries focus on the contextualisation clause; the extracted information is often wrong because from the moment the clause is compressed with the higher hierarchical level, inconsistencies are formed (except in the case where the additional level contains its own subject that allows generalising to the main case of subject, verb and object). 
Therefore, in cases that have structural rather than semantic problems, since the results would always be wrong and would have consistently lowered the percentage, we preferred to consider them exceptions.
The resulting percentages are shown in Table \ref{tab:tab-stats}.

As can be seen, the highest values are obtained over Subject and Verb. It may be argued that Objects tend to have lower HIT rates because they may be more complicated. The cases in which they are rather good, compared to the verb of the same item, on the other hand, are due to the presence of hierarchical levels that involve a greater number of sub-cases whose summation raises the value.

\begin{table}
\caption{Hits of the automated approach wrt the manual one}
\label{tab:tab-stats}
\begin{tabular}{|c|c|c|c|}
    \hline
    Art. & Sub-HIT & Verb-HIT  & Obj-HIT \\
    \hline
    7 & 100 \% & 100 \% & 91.6 \% \\
    \hline
    8 & 100 \% & 88.8 \% & 61.1 \% \\
    \hline
    9 & 100 \% & 100 \% & 43.3 \% \\
     \hline
    10 & 100 \% & 93.3 \% & 60 \% \\
    \hline
    11 & 95 \% & 90 \% & 45 \% \\
    \hline
    12 & 100 \% & 57.1 \% & 58.3 \% \\        
    \hline
     13 & 85.7 \% & 85.7 \% & 50 \% \\
    \hline
    14 & 90 \% & 100 \% & 50 \% \\
    \hline
    15 & 95 \% & 80 \% & 43.8 \% \\
    \hline
    16 & 100 \% & 87.5 \% & 66.6 \% \\
    \hline
    17 & - & - & - \\
    \hline
    18 & 100 \% & 50 \% & 75 \% \\
    \hline
    19 & 100 \% & 83.3 \% & 57.5 \% \\
    \hline
    20 & 75 \% & 75 \% & 100 \% \\
    \hline
    21 & 90 \% & 60 \% & 60 \% \\
    \hline
    22 & 100 \% & 100 \% & 50 \% \\
    \hline
    23 & 92.8 \% & 80.9 \% & 83.3 \% \\
    \hline
    24 & 80 \% & 80 \% & 40 \% \\
    \hline
    25 & 100 \% & 100 \% & 50 \% \\
    \hline
    26 & 100 \% & 33.3 \% & 33.3 \% \\
    \hline
    27 & 100 \% & 75 \% & 83.3 \% \\
    \hline
    28 & 100 \% & 91.6 \% & 86.6 \% \\
    \hline
    29 & 100 \% & 100 \% & 66.7 \% \\
    \hline
    30 & 100 \% & 100 \% & 100 \% \\
    \hline
     31 & 100 \% & 85.7 \% & 42.9 \% \\
    \hline
    32 & 100 \% & 75 \% & 67.1 \% \\
    \hline
    33 & 93.8 \% & 75 \% & 90\% \\ 
    \hline
    34 & 90 \% & 70 \% & 65\% \\
    \hline
    35 & 100 \% & 100 \% & 50\% \\    
    \hline
    36 & - & -  & - \\    
    \hline
    37 & 100 \% & 100 \% & 100\% \\
    \hline
    \textbf{Average} & {\textbf{ 96.1 \%}} &
    {\textbf{ 83.3 \%}} &  {\textbf{64.5 \%}} \\
    \hline
\end{tabular}

\end{table}

\section{Open challenges} \label{sec:open-ch}
As demonstrated by the statistical analysis, challenges can be encountered, with a certain relevance, through all cases of information extraction.

\paragraph{Identification of the Object}
The challenge of this type does not stem from the complexity of a single word: the pre-trained model is very good at identifying the English words encountered in the text. This can be demonstrated with a single and simple part of speech tagging.

On the contrary, it is difficult for the NLP library to extract the clauses and associate them with the correct place in the sentence.
An improvement here would be a breakthrough since one could remove the sentences that simply lengthen the sentence, without adding much additional information, and keep only those sentences that provide significant information.

Associating vast portions of text to either Complement (C) or Object (O) as it currently works limits extraction twice: even irrelevant details are associated causing an overly general tag and, at the same time, the specificity of the individual clauses is lost because many clauses are often tagged together. 

\paragraph{Identification of the style}
An article may begin with a lengthy contextualisation preamble. This is not infrequent in legal language and serves to contextualise the specific conditions for the application of the very measure that is described in the sequel of the article. This turns out to be a really challenging case for the automated approach. In fact, the POS are often extracted from the early parts in the sentence, with the result that they are obtained from the preamble rather than from the actual measure in this case, hence they are incorrect. 

\paragraph{Grammar issues}
Among these, we find the management of passive verbs. In their presence, the object is almost never identified. In fact, managing passive contexts is still a strong limitation for SpaCy.
Another problem is the use of pronouns. Of course, this is not exclusively connected to NLP but the association of nouns with their respective pronouns certainly remains an open problem, especially when referring to subjects. 

\section{Conclusions}\label{sec:concl}
This paper contributed to the essential interpretation of documents written in legal language by advancing an automated methodology for their grammatical tagging. The methodology leveraged the state of the art in the area of NLP, yet prescribing that the various outputs be validated by manual analysis. The methodology can be considered a general finding of this paper, namely it is arguably applicable to any target legal document following some preprocessing tailored to the specific features of the target.

The application of our methodology to the NIS 2 leads to the second finding, which consists of the first, albeit essential, structured interpretation of 
the entire directive. We have built all tables for the grammatical tagging of articles from 7 through to 37, showing the POS produced by leveraging the NLP libraries next to those produced manually by our own analysis and, finally, the relevant hit rates~\cite{repo}. Altogether, these tables may be leveraged in various ways in the future, for example as a reference even by those exercising the legal profession, and by ontologists aiming at the semantic representation of the directive.

A more technical finding is that forefront libraries SpaCy and ClausIE are very powerful and offer remarkable help. However, they are clearly put at stake by the complexities of the legal style. As a result, while their hit rates are very good over subjects and good over verbs, they are inadequate over objects. Therefore, our future work includes a more in-depth tailoring of these libraries, perhaps by re-training their models specifically for legal documents. We may expect the field of the automated processing of legal language to achieve valuable results in the near future.

\begin{acks}
We are grateful to Francesco Capparelli for  advice on EU legislation and to Mario Raciti for advice on semantic relations.
Gianpietro Castiglione acknowledges a studentship by Intrapresa S.r.l. and Italian ``Ministero dell’Università e della Ricerca'' (D.M. n. 352/2022).
\end{acks}

\bibliographystyle{ACM-Reference-Format}
\bibliography{biblio}

\begin{appendices}
\section*{Appendix}
\section*{Article 11 objects}
\subsection*{Item 11.1}
\subsubsection*{Manual}
    \begin{enumerate}[label=\textbf{N11.1.\arabic*}, leftmargin=1cm]
        \item \, 
        \begin{enumerate} [label=\textbf{N11.1.2.{\alph*}}]
            \item  a high level of availability of  their communication channels by avoiding single points of failure, and  shall  have  several  means  for  being  contacted  and  for  contacting  others  at  all times;  they  shall  clearly  specify  the communication channels and make them known to constituency and cooperative partners \label{n11-1-1a}
            \item located at secure sites\label{n11-1-1b}
            \item   with  an  appropriate  system  for  managing  and  routing  requests \label{n11-1-1c}
            \item the CSIRTs shall ensure the confidentiality and trustworthiness of  their operations \label{n11-1-1d}
            \item availability of  their services and their staff is trained appropriately \label{n11-1-1e}
            \item  redundant systems and backup working space to ensure continuity of their service \label{n11-1-1f}
        \end{enumerate}
    \end{enumerate}
    
\subsubsection*{ClausIE}
\begin{enumerate}[label=\textbf{S11.1.\arabic*}, leftmargin=1cm]
        \item \,
        \begin{enumerate} [label=\textbf{S11.1.2.{\alph*}}]
            \item a high level of availability of  their communication channels \label{s11-1-1a}
            \item NONE \label{s11-1-1b}
            \item NONE \label{s11-1-1c}
            \item the CSIRTs shall ensure the confidentiality and trustworthiness of  their operations \label{s11-1-1d}
            \item availability of  their services \label{s11-1-1e}
            \item NONE \label{s11-1-1f}
    \end{enumerate}
\end{enumerate}

\subsection*{Item 11.2}
\subsubsection*{Manual}
    \begin{enumerate}[label=\textbf{N11.2.\arabic*}, leftmargin=1cm]
        \item that  their  CSIRTs  jointly  have  the  technical  capabilities  necessary  to  carry  out  the  tasks referred  to  in  paragraph  3 \label{n11-2-1}
        \item that  sufficient  resources  are  allocated  to  their  CSIRTs  to  ensure adequate staffing levels for  the purpose of enabling the CSIRTs to develop their  technical capabilities \label{n11-2-2}
    \end{enumerate}

\subsubsection*{ClausIE}
    \begin{enumerate}[label=\textbf{S11.2.\arabic*}, leftmargin=1cm]
        \item that  their  CSIRTs  jointly  have  the  technical  capabilities  necessary  to  carry  out  the  tasks referred  to  in  paragraph  3 \label{s11-2-1}
        \item that  sufficient  resources  are  allocated  to  their  CSIRTs  to  ensure adequate staffing levels for  the purpose of enabling the CSIRTs to develop their  technical capabilities \label{s11-2-2}
    \end{enumerate}

\dots

\section*{Article 23 objects}
\subsection*{Item 23.1}
\subsubsection*{Manual}
    \begin{enumerate}[label=\textbf{N23.1.\arabic*}, leftmargin=1cm] 
         \item that essential and  important entities notify,  without undue  delay,  its  CSIRT or, where applicable,  its  competent  authority  in  accordance  with  paragraph  4  of  any  incident  that  has  a  significant  impact  on  the provision  of  their  services  as  referred  to  in  paragraph  3 \label{n23-1-1}
         \item the  recipients  of  their  services  of  significant  incidents  that  are  likely  to  adversely  affect  the provision  of  those  services \label{n23-1-2}
         \item that  those  entities  report,  inter  alia,  any  information  enabling the  CSIRT or,  where  applicable,  the  competent  authority  to  determine  any  cross-border  impact  of the  incident \label{n23-1-3}
         \item the notifying entity to increased liability \label{n23-1-4}
    \end{enumerate}
    
\subsubsection*{ClausIE}
    \begin{enumerate}[label=\textbf{S23.1.\arabic*}, leftmargin=1cm] 
        \item that essential and  important entities notify,  without undue delay,  its  CSIRT or, where applicable,  its  competent authority  in  accordance  with  paragraph  4  of  any  incident  that  has  a significant  impact  on  the provision  of  their  services  as  referred  to  in  paragraph  3 \label{s23-1-1}
        \item the  recipients  of  their  services  of  significant  incidents  that  are  likely  to  adversely  affect  the provision  of  those  services \label{s23-1-2}
        \item that  those  entities  report,  inter  alia,  any  information  enabling the  CSIRT or,  where  applicable,  the  competent  authority  to  determine  any  cross-border  impact  of the  incident \label{s23-1-3}
        \item the notifying entity to increased liability \label{s23-1-4}
    \end{enumerate}

\dots 

\subsection*{Item 23.8}
\subsubsection*{Manual}
    \begin{enumerate}[label=\textbf{N23.8.\arabic*}, leftmargin=1cm] 
        \item notifications received pursuant to paragraph 1 to the single points of contact of other affected Member States. \label{n23-8-1}
    \end{enumerate}
    
\subsubsection*{ClausIE}
    \begin{enumerate}[label=\textbf{S23.8.\arabic*}, leftmargin=1cm] 
        \item NONE \label{s23-8-1}
    \end{enumerate}

\subsection*{Item 23.9}
\subsubsection*{Manual}
    \begin{enumerate}[label=\textbf{N23.9.\arabic*}, leftmargin=1cm] 
        \item a  summary  report,  including anonymised and aggregated  data  on  significant  incidents,  incidents,  cyber  threats  and  near  misses  notified  in  accordance  with  paragraph  1 of  this Article  and  with  Article  3 \label{n23-9-1}
        \item technical  guidance  on  the  parameters  of  the  information  to  be  included  in  the  summary  report \label{n23-9-2} 
        \item the Cooperation Group and the CSIRTs network about its findings on notifications received every six months \label{n23-9-3}
    \end{enumerate}
    
\subsubsection*{ClausIE}
    \begin{enumerate}[label=\textbf{S23.9.\arabic*}, leftmargin=1cm] 
        \item a  summary  report,  including anonymised and aggregated  data  on  significant  incidents,  incidents,  cyber  threats  and  near  misses  notified  in  accordance  with  paragraph  1 of  this Article  and  with  Article  3 \label{s23-9-1}
        \item technical  guidance  on  the  parameters  of  the  information  to  be  included  in  the  summary  report \label{s23-9-2}
        \item the Cooperation Group and the CSIRTs network about its findings on notifications received every six months \label{s23-9-3}
    \end{enumerate}

\subsection*{Item 23.10}
\subsubsection*{Manual}
    \begin{enumerate}[label=\textbf{N23.10.\arabic*}, leftmargin=1cm] 
        \item  to  the  competent  authorities  under Directive  (EU)  2022/2557  information  about  significant  incidents,  incidents,  cyber  threats  and  near  misses  notified  in accordance  with  paragraph  1  of  this  Article  and  with  Article  30  by  entities  identified  as  critical  entities  under Directive (EU) 2022/2557 \label{n23-10-1}
    \end{enumerate}
    
\subsubsection*{ClausIE}
    \begin{enumerate}[label=\textbf{S23.10.\arabic*}, leftmargin=1cm] 
        \item NONE \label{s23-10-1}
    \end{enumerate}

\subsection*{Item 23.11}
\subsubsection*{Manual}
    \begin{enumerate}[label=\textbf{N23.11.\arabic*}, leftmargin=1cm ] 
        \item  implementing  acts  further  specifying  the  type  of  information,  the  format  and  the procedure  of  a  notification  submitted  pursuant  to  paragraph  1  of  this  Article  and  to  Article  30  and  of  a  communication submitted pursuant to paragraph 2 of  this Article \label{n23-11-1}
    \end{enumerate}
    
\subsubsection*{ClausIE}
    \begin{enumerate}[label=\textbf{S23.11.\arabic*}, leftmargin=1cm] 
        \item NONE \label{s23-11-1}
    \end{enumerate}
    
\end{appendices}
\end{document}